\pdfoutput=1
\documentclass[conference]{IEEEtran}

\usepackage[dvipsnames]{xcolor}
\usepackage{graphicx}
\usepackage{amsmath}
\usepackage{amssymb}
\usepackage{booktabs}
\usepackage[hyphens]{url}
\usepackage[numbers,sort&compress]{natbib}

\newcommand{\simname}{AlpaSim}
\newcommand{\vlaname}{Alpamayo-1.5}
\newcommand{\headname}{AlpaSafe}

\definecolor{statteal}{HTML}{2F7E74}

\newcommand{\regret}{\mathrm{regret}}
\newcommand{\Ltok}{\textsc{LastToken}}
\newcommand{\Lfull}{\textsc{Full}}
\newcommand{\Lnull}{\textsc{Null}}

\definecolor{linkblue}{rgb}{0.0,0.2,0.55}
\usepackage[breaklinks,colorlinks,allcolors=linkblue]{hyperref}
\hypersetup{
  pdftitle={AlpaSafe: Risk-Aware Trajectory Selection for Vision-Language-Action Models in Autonomous Driving},
  pdfauthor={Qiren Wang, Mingyu Guo, Weisong Shi},
}

\usepackage[capitalize]{cleveref}
\crefname{figure}{Fig.}{Figs.}
\Crefname{figure}{Fig.}{Figs.}
\crefformat{equation}{(#2#1#3)}
\Crefformat{equation}{Equation~(#2#1#3)}

\newcommand{\preprintnote}{This work has been submitted to the IEEE for possible publication. Copyright may be transferred without notice, after which this version may no longer be accessible.}

\makeatletter
\def\ps@preprintfirst{%
  \def\@oddhead{\parbox[b]{\textwidth}{\footnotesize\raggedright\preprintnote}}%
  \let\@evenhead\@oddhead
  \def\@oddfoot{\hfil\thepage\hfil}%
  \let\@evenfoot\@oddfoot}
\makeatother

\begin{document}

\title{AlpaSafe: Risk-Aware Trajectory Selection for\\Vision-Language-Action Models\\in Autonomous Driving}

\author{\IEEEauthorblockN{Qiren Wang, Mingyu Guo, and Weisong Shi}
\IEEEauthorblockA{Connected and Autonomous Research Laboratory\\
University of Delaware, Newark, USA\\
\{qirenw, myuguo, weisong\}@udel.edu}}

\bstctlcite{IEEEexample:BSTcontrol}

\maketitle
\thispagestyle{preprintfirst}

\begin{abstract}
Reasoning-based driving vision-language-action (VLA) models propose multiple candidate trajectories at each decision, yet only one is executed.
Standard evaluation scores the best candidate, assuming an oracle selector that deployment lacks, and leaves the risk of the executed candidate unexamined.
Existing online evaluators typically close this gap with a purpose-trained scene or world model, often rolling out each candidate through it at inference time.
We introduce \headname{}, a risk-aware trajectory selector that reuses the scene representation a frozen driving VLA computes while proposing its candidates.
\headname{} reads the last hidden token of the VLA's prompt prefill. A lightweight head, trained on the closed-loop simulation outcomes of the candidates, fuses this token with each candidate trajectory and predicts a relative score and collision/offroad logits.
A fixed rule removes candidates dominated on the two risk logits and selects the highest-scoring candidate among those remaining, adding neither rollouts nor VLA training.
Experiments on 525 \simname{} scenes show that \headname{} outperforms Alpamayo's default first-sample rule and uniform random selection in selected regret and top-1 agreement, and reduces the collision-at-fault or offroad rate from 6.41\% and 6.10\% to 4.93\%.
The selector has 1.7M parameters and takes 0.75\,ms per decision.
\end{abstract}


\section{Introduction}
\label{sec:intro}

\begin{figure*}[t]
  \centering
  \includegraphics[width=\textwidth]{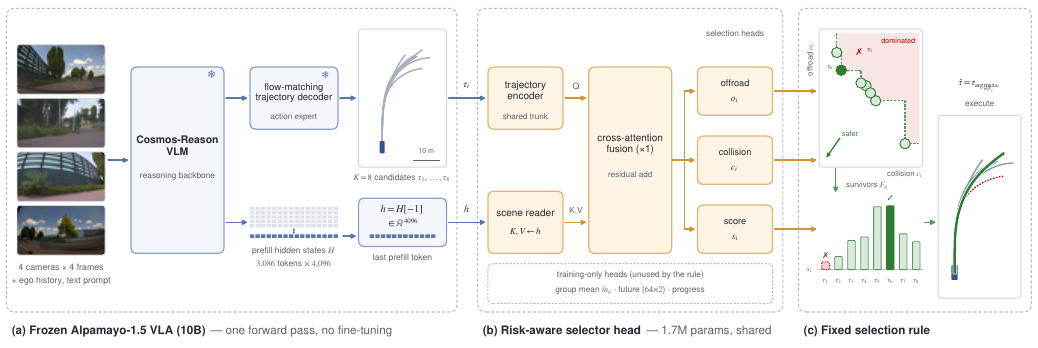}
  \caption{\headname{} overview. (a) One forward pass of the frozen \vlaname{} generator yields the $K{=}8$ candidate trajectories and the prompt-prefill hidden states $H$; the selector reads only the last token $h=H[-1]$. (b) The selector head encodes each candidate, fuses it with $h$ through one cross-attention block with a residual connection, and outputs a bounded relative score $s_i$ and collision/offroad logits $(c_i, o_i)$; the group-mean, future-trajectory, and progress heads supervise training only. (c) Candidates dominated on the two risk logits are removed, and the highest $s_i$ among the survivors is executed. Camera frames, candidates, scores, and logits are recorded data from one decision group of the evaluation pool (\Ltok{}).}
  \label{fig:arch}
\end{figure*}

Reasoning-based vision-language-action (VLA) models are increasingly used as end-to-end planners for autonomous driving~\cite{tian2024drivevlm,hwang2024emma,alpamayo2025}. Given a scene and a driving prompt, these models can generate multiple plausible trajectory candidates rather than committing to a single future. This creates a trajectory selection problem, because only one candidate can be executed.
Proposal-based driving systems therefore learn trajectory scorers or selectors to rank the generated candidates~\cite{kirby2026drivor}. Best-of-$K$ metrics such as minADE@$K$, however, select the best candidate in hindsight and thus mainly measure whether the generated set contains a good trajectory. At deployment, the system must instead determine which candidate is most likely to produce the best closed-loop outcome before any candidate is executed.

Existing online trajectory evaluators commonly address this problem through explicit imagination. WoTE~\cite{li2025wote}, for example, conditions a BEV world model on each candidate, predicts candidate-specific future BEV states, and evaluates the trajectory from those imagined outcomes. This makes future consequences explicit, but it requires a purpose-trained world model and additional candidate-conditioned prediction at inference time.
Yet a large reasoning-based VLA has already integrated the observed scene, driving intent, and interaction context while generating its candidate trajectories. It remains unknown whether the hidden representation produced during this process contains enough scene-specific information to evaluate the candidates without explicitly predicting their futures.

To answer this question, we introduce \headname{} (\cref{fig:arch}), a risk-aware trajectory selector for reasoning-based driving VLAs. \headname{} captures the final-layer hidden states of the frozen VLA's prompt prefill during the same forward pass that generates the candidates and reads only the last token, a byproduct of generation that requires no additional VLA computation. A lightweight head then scores each candidate against this token and filters the candidates on predicted collision and offroad risk.

To isolate selection quality, we hold the candidate pool and its official closed-loop outcome labels from the \simname{} simulator fixed across methods, and learned-selector predictions are out of fold at the scene level. Matched ablations that replace the last token with a zero token or with the full prefill sequence test whether the reused representation matters, and uncertainty is reported with paired scene-bootstrap intervals. The selector has 1.7M parameters and runs its full selection path in 0.75\,ms per eight-candidate decision on one GPU, whereas one closed-loop candidate rollout in the same simulator takes about 111\,s.

The main contributions of this work are summarized as follows:
\begin{itemize}
  \item We propose \headname{}, a risk-aware trajectory selection framework for reasoning-based driving VLAs that reuses the last prompt-prefill hidden token of the frozen generator, captured in the same forward pass that produces the candidates, and requires neither VLA training nor inference-time rollouts.
  \item We design a risk-aware selector head, a lightweight action-conditioned module that scores each candidate trajectory with a relative outcome score and collision/offroad risk logits, removes the candidates dominated on the two risk logits, and selects the highest-scoring trajectory among those remaining.
  \item Experimental results on \simname{} scenes demonstrate that \headname{} outperforms Alpamayo's default first-sample rule and uniform random selection in selected regret and top-1 agreement, and reduces the collision-at-fault or offroad rate from 6.41\% and 6.10\% to 4.93\%, at only 0.75\,ms per decision.
\end{itemize}

\section{Related Work}
\label{sec:related}

Driving scenes often admit several valid maneuvers, motivating candidate-based formulations that preserve multiple plausible trajectories rather than regress a single averaged future~\cite{chai2019multipath,phanminh2020covernet}.
End-to-end planners encode the current observation using BEV, vectorized, or object-level scene representations~\cite{hu2023uniad,jiang2023vad,chen2024vadv2,renz2023plant}, while driving foundation models combine vision-language features with generative trajectory decoders~\cite{shao2024lmdrive,tian2024drivevlm,hwang2024emma,alpamayo2025}.
Once multiple ego trajectories are generated, however, planning becomes a selection problem.

One family of methods scores each candidate directly from its geometry and the current scene, without predicting a candidate-conditioned future at inference time.
Early neural motion planners learn costs over sampled feasible trajectories~\cite{zeng2019nmp}; subsequent approaches distill metric-specific teachers or learn generalized trajectory values and metric sub-scores~\cite{li2024hydramdp,li2025gtrs,yao2026drivesuprim,kirby2026drivor,ang2026clover}.
TOAD further treats a learned scorer as the objective for test-time trajectory optimization~\cite{xu2026toad}.
A central distinction among these methods is the scene evidence exposed to the scorer.
DrivoR uses multi-camera register tokens shared by its generation and scoring decoders, while recent studies show that intermediate VLA states can support downstream driving decisions and trajectory refinement~\cite{ling2026blue,babu2026planningtoken,yang2026lwdrive}.
\headname{} adopts this representation-reuse setting, combining every candidate with a shared representation extracted from a frozen driving VLA and learning a per-trajectory score.

A second family evaluates a candidate only after constructing evidence about its possible consequences.
PDM rolls out kinematic proposals against simplified agent forecasts and ranks them using hand-designed rules~\cite{dauner2023pdm}.
DTPP replaces this fixed evaluation with jointly learned ego-conditioned prediction and context-aware cost evaluation for tree-structured policy candidates~\cite{huang2024dtpp}.
Several world-model methods construct richer candidate-conditioned futures: WoTE predicts dense future BEV states, SafeDrive represents safety-critical future entities in a sparse world, and World4Drive uses an intention-aware latent world model to support multimodal trajectory selection~\cite{li2025wote,kim2026safedrive,zheng2025world4drive}.
Foundation-model approaches instead employ a generative VLM reward model or a VLA critic to evaluate and refine proposed trajectories~\cite{chen2026drivereward,yang2026criticvla}.

The principal distinction is therefore whether selection requires constructing a candidate-conditioned future world at decision time, not whether future prediction is used as a training signal.
WPT removes the inference-time cost by distilling world-model guidance into a lightweight policy and discarding the world model at deployment~\cite{jiang2026wpt}.
\headname{} uses consequence supervision to train a per-trajectory scorer and event-risk heads rather than the trajectory-generation policy: each candidate is evaluated directly against the frozen VLA scene representation, and no candidate-conditioned future world is predicted at inference time.

\section{Methodology}
\label{sec:method}

\Cref{fig:arch} gives an overview of \headname{}: one forward pass of the frozen VLA yields both the $K$ candidate trajectories and the prefill hidden states, and the selector pairs the last token of those states with each candidate.

\subsection{Problem Setting}
\label{sec:setting}

We work in \simname, a high-fidelity closed-loop driving simulator in the tradition of public closed-loop benchmarks~\cite{caesar2021nuplan,dauner2024navsim,jia2024bench2drive}, whose official scene score $y\in[0,1]$ aggregates safety, compliance, and progress into a single scalar per rollout.
Each scene contributes three decision groups at staggered decision times.
For each group, the frozen 10B-parameter generator \vlaname{}, a reasoning VLM coupled to a trajectory decoder (\cref{fig:arch}a), samples $K{=}8$ candidate trajectories from one prompt prefill. Every candidate is then executed in closed loop and receives its own official score, giving 24 scoring rollouts per scene.
In our data the official score equals zero when collision-at-fault or offroad occurs and otherwise equals the simulator's \texttt{progress\_score}. Each rollout also carries separate collision-at-fault and offroad event labels.

Given a group $g$ with candidates $\tau_1,\dots,\tau_K$ and scores $y(\tau_k)$, a selector $\pi$ commits to $\hat\tau=\pi(g)$.
We evaluate two primary quantities:
selected regret, $\regret(g) = \max_k y(\tau_k) - y(\hat\tau)$, the selected candidate's gap from the oracle outcome, for which lower is better, and
tie-aware top-1 agreement, $\mathbf{1}[\,y(\hat\tau) \ge \max_k y(\tau_k) - \epsilon\,]$ with $\epsilon=10^{-6}$, for which higher is better.
Both are averaged over the groups of a scene and then equally over scenes (scene-macro mean).

\subsection{Same-Prefill Representation Extraction}
\label{sec:capture}

\begin{figure}[t]
  \centering
  \includegraphics[width=\linewidth]{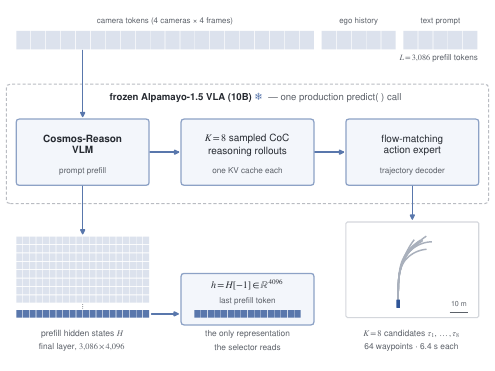}
  \caption{Same-prefill representation extraction. One production \texttt{predict()} call per decision group runs one prompt prefill (camera, ego-history, and text tokens) through the frozen generator and yields two products of that forward pass: the $K{=}8$ candidates, decoded by the flow-matching action expert from $K$ sampled chain-of-causation (CoC) reasoning traces, and the final-layer hidden states $H$ of the 3{,}086-token prefill. The selector consumes only the last row $h=H[-1]$. The candidates shown are the recorded outputs of the decision group in \cref{fig:arch}.}
  \label{fig:capture}
\end{figure}

A representation-reuse claim is confounded if the representation could come from a different forward pass than the candidates, for example a different context window, different preprocessing, or a silently re-run prefill.
We eliminate this by construction.
A capture hook inside the production \texttt{predict()} path records the final transformer layer's post-normalization hidden states of the prompt prefill, $H\in\mathbb{R}^{3086\times4096}$, during the same call that samples the $K$ candidates (\cref{fig:capture}); the prefill length is the same for every group.
An exactly-one-prefill assertion makes a second prefill in the call fail the capture rather than silently substitute context.
The scene vector used at deployment is the last row, $h = H[-1]\in\mathbb{R}^{4096}$.
Capture-time SHA-256 hashes of both tensors establish bit-exact provenance, with $h \equiv H[-1]$ verified for every group, and every downstream artifact, from training tables to locked predictions and analysis records, carries these hashes forward.
Candidates and representation are therefore co-produced, and the closed-loop scores remain linked to the corresponding candidates.

\subsection{Risk-Aware Selector Head}
\label{sec:head}

\begin{figure}[t]
  \centering
  \includegraphics[width=\linewidth]{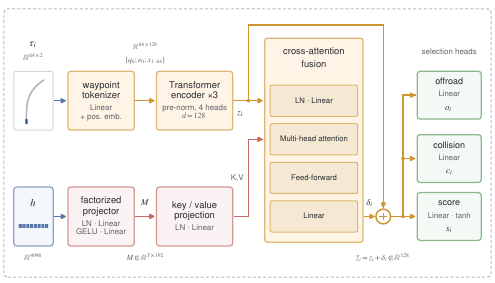}
  \caption{Expanded view of the selector head, panel (b) of \cref{fig:arch}. Blue marks the products of the frozen generator ($\tau_i$, $h$); the other blocks are trained: yellow runs per candidate, red is the scene reader, computed once per group, and green marks the heads read by the rule of panel (c). The scene memory $M$ has one row for \Ltok{} and 3{,}086 rows for \Lfull{}, and $\oplus$ denotes a residual add.}
  \label{fig:head}
\end{figure}

The \headname{} head has 1{,}708{,}871 trainable parameters and scores each candidate independently; the frozen VLA receives no gradients (\cref{fig:arch}).
There is no cross-candidate attention, so scores are $K$-invariant and permutation-equivariant by construction. We verify that off-diagonal cross-candidate gradients are exactly zero and that scores are bit-identical whether a candidate is scored alone or in a batch.
\Cref{fig:head} expands panel (b) of \cref{fig:arch}.
We compare three scene-input variants of this head: \Ltok{}, the proposed selector, reads only $h$; \Lfull{} reads all rows of $H$; and \Lnull{} reads a zero vector.
All widths below were fixed before any evaluation-pool training and are shared by the three variants.
Throughout, vectors are columns, a token sequence is a matrix with one token per row, $\operatorname{LN}$ denotes layer normalization, and every linear map carries a bias that we omit from the notation.

\paragraph{Trajectory encoder}
A candidate $\tau_i$ consists of 64 normalized waypoints $\tau_{i,t}\in\mathbb{R}^{2}$.
Each waypoint is embedded by a shared linear map with a learned positional embedding, a learned read-out token $q_0$ and a constant placeholder token $e_0$ are prepended, and the 66 tokens are encoded jointly:
\begin{equation}
\begin{gathered}
  x_{i,t} = W_{\mathrm{tok}}\,\tau_{i,t}+e^{\mathrm{pos}}_{t},\qquad t=1,\dots,64,\\
  [\,z_i;\,\cdot\,;\,Z_i\,] = \operatorname{Enc}\big([\,q_0;\,e_0;\,x_{i,1};\dots;x_{i,64}\,]\big).
\end{gathered}
\label{eq:enc}
\end{equation}
Here $\operatorname{Enc}$ is three pre-norm Transformer encoder layers (4 heads, width 128, feed-forward 256, GELU, dropout 0.1) followed by a final LayerNorm. Its output at the read-out position is the candidate latent $z_i\in\mathbb{R}^{128}$, its outputs at the 64 waypoint positions are the waypoint states $Z_i\in\mathbb{R}^{64\times128}$, and its output at the placeholder position is discarded.
The placeholder receives a fixed vector in place of a scene token, so the encoder is scene-free and the scene reaches a candidate only through the fusion block.

\paragraph{Scene reader}
Once per group, the scene input is projected to a memory through a width-64 bottleneck.
Let $h_1,\dots,h_T\in\mathbb{R}^{4096}$ be the visible rows of $H$: the single row $h$ for \Ltok{} ($T{=}1$), all $T{=}3{,}086$ rows for \Lfull{}, and one zero row for \Lnull{}. Then
\begin{equation}
  m_t = W_{\mathrm{up}}\operatorname{GELU}\!\big(W_{\mathrm{dn}}\operatorname{LN}(h_t)\big),\qquad t=1,\dots,T,
  \label{eq:memory}
\end{equation}
with $W_{\mathrm{dn}}\in\mathbb{R}^{64\times4096}$ and $W_{\mathrm{up}}\in\mathbb{R}^{192\times64}$, and the rows $m_t$ form the memory $M=[\,m_1;\dots;m_T\,]\in\mathbb{R}^{T\times192}$.

\paragraph{Fusion}
A single pre-norm cross-attention block with residual connections lets each candidate query the memory, and the result is added back to the candidate's encoder states:
\begin{equation}
\begin{aligned}
  u_i &= W_{\mathrm{in}}\,z_i,\\
  u'_i &= u_i+\operatorname{MHA}\big(\operatorname{LN}(u_i),\,\operatorname{LN}(M)\big),\\
  \delta_i &= W_{\mathrm{out}}\big(u'_i+\operatorname{FFN}(\operatorname{LN}(u'_i))\big),\\
  \tilde z_i &= z_i+\delta_i,\\
  \tilde Z_i &= Z_i+\mathbf{1}\,\delta_i^{\top}.
\end{aligned}
\label{eq:fusion}
\end{equation}
Here $W_{\mathrm{in}}\in\mathbb{R}^{192\times128}$ lifts the latent to the reader width, $\operatorname{MHA}(a,B)$ is six-head scaled dot-product attention (head width 32) of a query $a$ over the rows of $B$ with linear query, key, value, and output projections, $\operatorname{FFN}$ is a $192\!\to\!384\!\to\!192$ GELU network, and $W_{\mathrm{out}}\in\mathbb{R}^{128\times192}$ maps back to $\delta_i\in\mathbb{R}^{128}$. The all-ones vector $\mathbf{1}\in\mathbb{R}^{64}$ adds $\delta_i$ to every waypoint state.
For \Ltok{} the memory has one row, so the attention weights are identically one and the block reduces to adding one projection of $h$ to every candidate query $u_i$, followed by the residual FFN; the same graph over $T{=}3{,}086$ rows is the \Lfull{} ablation, which keeps the variants parameter-matched.

\paragraph{Output heads}
The three selection-path heads are linear in the fused latent,
\begin{equation}
\begin{aligned}
  \hat r_i &= w_r^{\top}\tilde z_i,\\
  c_i &= w_c^{\top}\tilde z_i,\\
  o_i &= w_o^{\top}\tilde z_i,
\end{aligned}
  \label{eq:heads}
\end{equation}
giving the per-candidate score residual $\hat r_i$, whose head is zero-initialized, and the collision and offroad logits $c_i$ and $o_i$.
Three further outputs are used only during training: the progress $\hat p_i=w_p^{\top}\tilde z_i$, the future correction $\Delta_i=\tilde Z_i W_{\mathrm{fut}}^{\top}\in\mathbb{R}^{64\times2}$ from the waypoint states, and the group mean $\hat m_g=w_m^{\top}\tanh\!\big(W_{\mathrm{grp}}\operatorname{LN}(\bar h)\big)$ from the mean $\bar h=\frac{1}{T}\sum_t h_t$ of the visible rows ($\bar h=h$ for \Ltok{}), which never sees a candidate.
The parameters split into 0.94M for the trajectory encoder (0.53M of them in the placeholder token's adapter), 0.63M for the scene reader and fusion block, 0.14M for the group-mean branch, and 774 for the five linear output heads.

\paragraph{Score decomposition and selection rule}
The official score mixes the group's overall difficulty with each candidate's relative quality, so we decompose the score prediction as
\begin{equation}
\begin{aligned}
  \hat y_i &= \hat m_g + s_i,\\
  s_i &= S\tanh\!\big(\hat r_i/S\big),
\end{aligned}
\label{eq:decomp}
\end{equation}
where $\hat m_g$ is the group-mean prediction, $s_i$ is the bounded candidate-relative score obtained from the residual $\hat r_i$, and the bound $S{=}0.67$, frozen from development-set residual statistics, keeps the residual inside the empirically observed spread.
Candidate $j$ dominates $i$ when $c_j\le c_i$ and $o_j\le o_i$, with at least one strict inequality. Let $F_g$ contain the candidates not dominated by another candidate in the group. The selection rule is
\begin{equation}
  i^\star=\operatorname*{arg\,max}_{i\in F_g}\, s_i,
  \label{eq:selection}
\end{equation}
with exact score ties resolved by the smallest candidate index, and the selected trajectory is $\hat\tau=\tau_{i^\star}$. $F_g$ is never empty, so exactly one candidate is selected without a fallback. The rule compares raw logits, uses no risk threshold or fusion weight, and ignores the group-mean offset, which is constant within a group. The risk heads are trained with weighted binary cross-entropy, so their sigmoid outputs are not treated as calibrated probabilities, and non-dominance is a comparison among predicted risks, not an absolute safety guarantee. The same rule is applied to all three variants.

\subsection{Training Objective}
\label{sec:objectives}

The three variants are trained with the same objective.
For a group $g$, write $y_i=y(\tau_i)$ for the official score of candidate $i$ and $\bar y_g=\frac{1}{K}\sum_i y_i$ for the group mean.
The objective is a fixed weighted sum of seven terms,
\begin{equation}
\begin{split}
  L \;=\; & 1.0\,L_{\text{rank}} + 0.5\,L_{\text{centered}} + 0.1\,L_{\text{group-mean}} \\
  & + 0.25\,L_{\text{future}} + 0.1\,L_{\text{progress}} \\
  & + 0.1\,L_{\text{collision}} + 0.1\,L_{\text{offroad}},
\end{split}
\label{eq:objective}
\end{equation}
whose coefficients, target scales, and checkpoint rule were fixed on the development set before any evaluation-pool training.
Regression terms use the SmoothL1 loss $\rho_\beta(d)=d^2/(2\beta)$ for $|d|<\beta$ and $|d|-\beta/2$ otherwise.

\paragraph{Score terms}
The first three terms supervise the decomposed score of \cref{eq:decomp}.
$L_{\text{rank}}$ is a margin-weighted pairwise logistic loss on the bounded relative scores $s_i$.
Let $\mathcal{P}$ be the candidate pairs $(i,j)$ of the groups in a mini-batch with $|y_i-y_j|>10^{-6}$; exact ties are excluded. Then
\begin{equation}
  L_{\text{rank}} = \sum_{(i,j)\in\mathcal{P}} \tilde w_{ij}\,
  \operatorname{softplus}\!\big(-\operatorname{sgn}(y_i-y_j)\,(s_i-s_j)\big),
  \label{eq:lrank}
\end{equation}
with $\tilde w_{ij}=|y_i-y_j|/\sum_{\mathcal{P}}|y_i-y_j|$, so pairs with a large outcome gap dominate and any per-group offset cancels.
Ranking alone leaves the scale of $s_i$ free, so $L_{\text{centered}}=\frac{1}{K}\sum_i \rho_{\beta_c}\big(s_i-(y_i-\bar y_g)\big)$ regresses $s_i$ onto the group-centered official score, and $L_{\text{group-mean}}=\rho_{\beta_m}(\hat m_g-\bar y_g)$ supervises the candidate-independent group-mean head, so that between-group difficulty is absorbed by $\hat m_g$ rather than by $s_i$.
The knees match the empirical scale of each target: $\beta_c=0.075$ is the mean within-group standard deviation of the official score and $\beta_m=0.243$ its between-group standard deviation.
Typical errors then fall in the linear regime of $\rho_\beta$, so the per-candidate gradients of $L_{\text{rank}}$ and $L_{\text{centered}}$ have matched magnitude at initialization and the coefficients in \cref{eq:objective} approximate the realized gradient ratios; with the default $\beta=1$ the realized rank-to-centered ratio would be more than ten times the stated $2{:}1$.

\paragraph{Future, progress, and risk terms}
The last four terms supervise the heads of \cref{sec:head} from the executed rollout of each candidate.
The future head predicts a $64\times2$ correction $\Delta_i$ to the planned waypoints, and $L_{\text{future}}$ is the elementwise $\rho_{1}$ between $\tau_i+\Delta_i$ and the executed ego positions over the 64 simulation steps after the decision, in the ego frame at decision time, averaged over the valid steps of the $K$ candidates.
$L_{\text{progress}}=\frac{1}{K}\sum_i\rho_1(\hat p_i-p_i)$ regresses the simulator's clipped relative route progress $p_i\in[0,1]$ of the rollout.
$L_{\text{collision}}$ and $L_{\text{offroad}}$ are binary cross-entropy with logits on the per-candidate collision-at-fault and offroad labels, with the positive class weighted by $n_{\text{neg}}/n_{\text{pos}}$ of the fold's training candidates (about 43 and 24, as 289 and 496 of the 12{,}600 rollouts are positive).

\paragraph{Optimization}
Each variant is trained with AdamW (learning rate $3\times10^{-4}$, weight decay $10^{-4}$) and gradient clipping at $1.0$, two groups per mini-batch with four-step gradient accumulation, for at most 100 epochs with patience 20.
Pairs in \cref{eq:lrank} are pooled across the groups of a mini-batch; every other term is averaged over them.
The checkpoint is chosen on the inner-validation split only: among the epochs whose pairwise ranking accuracy is within $0.01$ of the running best, the one with the lowest selected regret is kept, with ties broken by lower best-versus-second inversion rate, higher top-1 agreement, higher best-versus-rest pairwise accuracy, and the earlier epoch; early stopping uses the same key.

\subsection{Evaluation Protocol}
\label{sec:protocol}

Training and evaluation use five-fold cross-validation with one scene-level partition shared by all learned selectors; each fold has 357 training, 63 inner-validation, and 105 test scenes. Every scene is scored by the model that held it out.
The selection rule in \cref{eq:selection} was adopted after earlier outcome analyses on these scenes. It is applied to the stored out-of-fold relative scores and collision/offroad logits, with model weights and predictions unchanged. The common-pool comparison is therefore exploratory rather than an independent confirmatory test.
Uncertainty estimates use a paired scene percentile bootstrap with $10{,}000$ resamples that keeps all groups of a sampled scene together. The 95\% intervals are pointwise, without multiple-comparison correction, and we call a difference supported when its interval excludes zero.

\section{Experiments}
\label{sec:exp}

\begin{table}[t]
  \centering\small\setlength{\tabcolsep}{3.5pt}
  \caption{Results on all 525 scenes and 1{,}575 groups, with $K{=}8$. Score is the selected candidate's true official score, and all values are scene-macro means.}
  \label{tab:main-abs}
  \begin{tabular}{@{}lccc@{}}
    \toprule Selector & Regret $\downarrow$ & Top-1 $\uparrow$ & Score $\uparrow$ \\
    \midrule
    Alpamayo default & $0.0611$ & $0.239$ & $0.5635$ \\
    Uniform random & $0.0586$ & $0.238$ & $0.5660$ \\
    \Ltok{} (ours) & $0.0512$ & $0.348$ & $0.5734$ \\
    \bottomrule
  \end{tabular}
\end{table}
\begin{figure}[t]
  \centering\includegraphics[width=\linewidth]{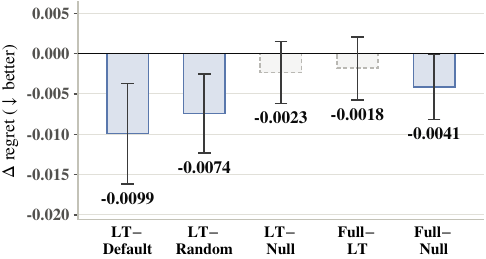}
  \caption{Paired regret differences on 525 shared scenes. Bars show mean differences and whiskers show 95\% scene-bootstrap intervals. Blue bars with a solid outline mark intervals excluding zero, all favoring the first-named rule; gray bars with a dashed outline mark intervals including zero. Negative values favor the first-named rule. LT denotes \Ltok{}.}
  \label{fig:results-regret}
  \vspace{\floatsep}
  \centering\includegraphics[width=\linewidth]{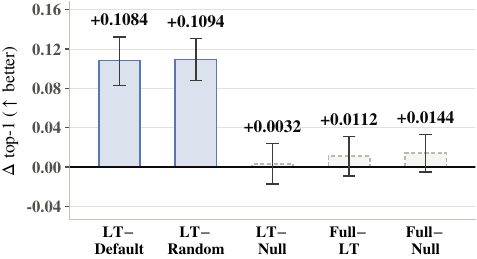}
  \caption{Paired top-1 differences on the same 525 scenes. Positive values favor the first-named rule; other conventions follow \cref{fig:results-regret}.}
  \label{fig:results-top1}
\end{figure}
\begin{figure*}[t]
  \centering\includegraphics[width=\textwidth]{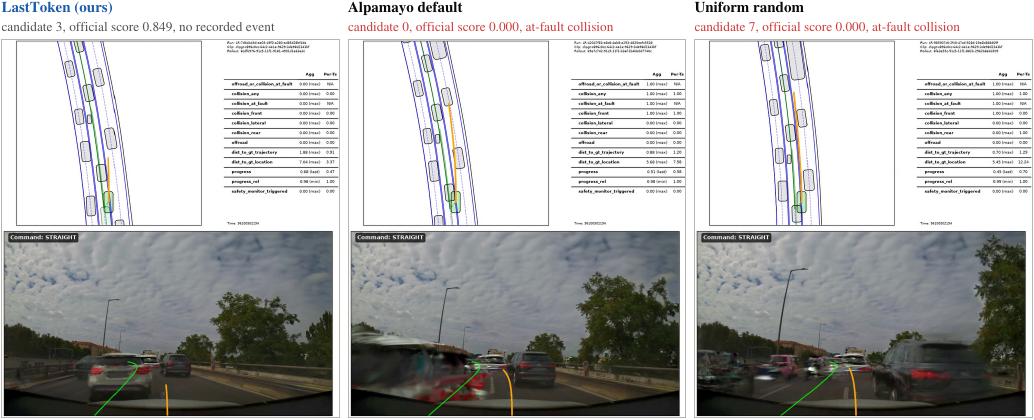}
  \caption{Recorded outcomes for one evaluation-pool decision group, 3.8\,s after the decision. Each panel shows the top-down view with metric summary above the front camera. Orange (light gray in grayscale print) marks the executed trajectory and green (mid gray) the navigation route and, in the top-down view, the recorded human drive. \Ltok{} selects the only one of the eight candidates without a recorded safety failure. The groups in \cref{fig:case-main,fig:case-null,fig:case-full} were chosen retrospectively for camera legibility and show individual selections, not average performance.}
  \label{fig:case-main}
\end{figure*}
\begin{figure*}[t]
  \centering\includegraphics[width=\textwidth]{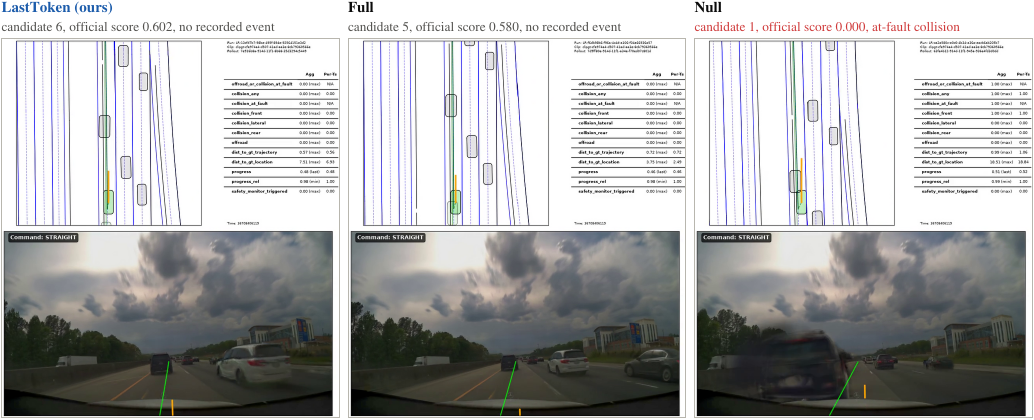}
  \caption{Recorded outcomes for the scene-information ablation, 6.1\,s after the decision, with conventions as in \cref{fig:case-main}. \Lnull{} has selected the only colliding candidate of its group, whereas \Ltok{} and \Lfull{} selected event-free candidates.}
  \label{fig:case-null}
\end{figure*}
\begin{figure*}[t]
  \centering\includegraphics[width=\textwidth]{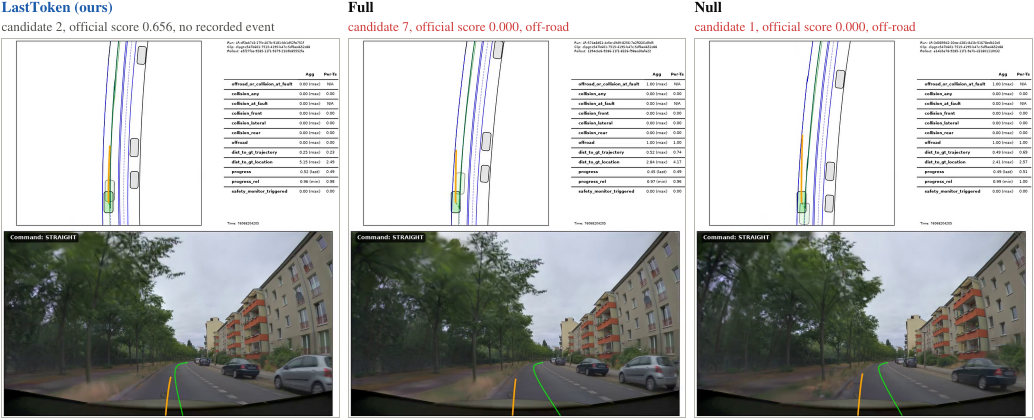}
  \caption{Recorded outcomes for the last-token versus full-sequence comparison, 5.2\,s after the decision, with conventions as in \cref{fig:case-main}. The candidates selected by \Lfull{} and \Lnull{} have left the road, whereas the \Ltok{} selection has not.}
  \label{fig:case-full}
\end{figure*}

\subsection{Experimental Setup}
\label{sec:setup}

Every selection rule and learned ablation is compared on the same evaluation pool of 525 scenes, 1{,}575 decision groups, and 12{,}600 scored candidates, with identical candidates and official outcome labels.
A separate 122-scene development set was used for architecture and objective design, and a further 35-scene reserve remains sealed and unused.

We report selected regret, tie-aware top-1 agreement, mean official score, and event rates at $K{=}8$.
The event rates are the collision rate, the offroad rate, and the either-event rate, the scene-macro proportions of selections that end in collision-at-fault, in offroad, or in either.
The statistical unit is the scene ($n{=}525$), not the group or the candidate.
Rollouts that end in collision or offroad are valid zero-score outcomes and are retained.

\subsection{Comparison with Default and Random Selection}
\label{sec:res-selection}

Default executes candidate index zero, as in the reference Alpamayo driver.
Uniform random selects among the same eight candidates with a fixed pseudo-random draw per group, repeated three times.
The three draws select from the same cached candidates and are not independently sampled test sets; we average their metrics within each scene before scene-level aggregation.
\Cref{tab:main-abs} reports the three selection rules.
\Ltok{} lowers regret by $0.0099$ relative to default and by $0.0074$ relative to random, and raises top-1 agreement by $0.11$ relative to each.
All four differences are supported; \cref{fig:results-regret,fig:results-top1} show them with 95\% scene-bootstrap intervals, together with the ablation differences discussed below.
This comparison measures selection from fixed candidates with cached closed-loop labels; it does not evaluate repeated online replanning under the learned selector.
\Cref{fig:case-main} shows one group from this pool in which \Ltok{} executes the only candidate without a recorded safety failure, while the default and random selections each end in an at-fault collision.

\subsection{Safety Outcomes}
\label{sec:res-safety}
The either-event rate is 4.93\% for \Ltok{}, 6.41\% for default, and 6.10\% for random.
The differences, $-1.48$ percentage points versus default and $-1.16$ points versus random, are both supported.
\Ltok{}'s separate collision and offroad rates are 1.57\% and 3.37\%; taken separately, the collision decrease versus default and the offroad decrease versus random are not supported.

Official-score optimization and event-rate minimization are different objectives even though an event zeros the official score. Filtering can remove a candidate with a higher relative score, so event rates must be read together with regret and official score. Across all \Ltok{} selections, the mean retained-set size is $2.212$ of the eight candidates, and no final-score ties occur.

\subsection{Matched Ablations: Scene Information}
\label{sec:res-signal}\label{sec:res-scale}

\begin{table}[t]
  \centering\small\setlength{\tabcolsep}{3.5pt}
  \caption{Matched scene-input ablations, with scenes, groups, and columns as in \cref{tab:main-abs}. Only the scene input differs: the last token for \Ltok{}, one zero token for \Lnull{}, and all 3{,}086 tokens for \Lfull{}.}
  \label{tab:ablation-abs}
  \begin{tabular}{@{}lccc@{}}
    \toprule Selector & Regret $\downarrow$ & Top-1 $\uparrow$ & Score $\uparrow$ \\
    \midrule
    \Ltok{} (ours) & $0.0512$ & $0.348$ & $0.5734$ \\
    \Lnull{} & $0.0536$ & $0.345$ & $0.5710$ \\
    \Lfull{} & $0.0494$ & $0.359$ & $0.5752$ \\
    \bottomrule
  \end{tabular}
\end{table}

\Cref{tab:ablation-abs} reports the three matched selectors.
\Lnull{} replaces the scene representation by one zero token and is otherwise identical to \Ltok{}, so the comparison isolates the scene information available to the selector.
\Ltok{} minus \Lnull{} gives regret $-0.0023$ and top-1 $+0.003$; neither difference is supported (\cref{fig:results-regret,fig:results-top1}).
The event rates do differ: with the last token, the collision rate is lower by $0.66$ percentage points and the either-event rate by $0.72$ points, and both differences are supported.
\Cref{fig:case-null} shows a group in which \Lnull{} selects the only colliding candidate of the eight while \Ltok{} and \Lfull{} select event-free candidates.

\subsection{Last Token versus Full Sequence}
\label{sec:res-onetoken}

\Lfull{} reads all 3{,}086 prefill tokens and \Ltok{} only the final one.
\Lfull{} minus \Ltok{} gives regret $-0.0018$ and top-1 $+0.011$; neither difference is supported, and the intervals do not establish equivalence either.
\Lfull{} has a higher collision rate, by $0.57$ percentage points (supported), whereas the either-event difference of $+0.42$ points is not supported.
The remaining comparison in \cref{fig:results-regret,fig:results-top1}, \Lfull{} minus \Lnull{}, gives regret $-0.0041$ and top-1 $+0.014$; only the regret difference is supported.
\Cref{fig:case-full} shows a group in which the candidates selected by \Lfull{} and \Lnull{} leave the road while \Ltok{} selects an event-free candidate.

\subsection{Deployment Cost}
\label{sec:res-eff}

\begin{table}[t]
  \centering\small\setlength{\tabcolsep}{4pt}
  \caption{Cost of one eight-candidate decision: closed-loop simulation of every candidate versus the \Ltok{} selection path. Both columns were measured on the same NVIDIA RTX PRO 6000 GPU. Simulation entries use the median of 24 candidate rollouts (104.5\,s) plus official scoring (6.8\,s); the selector entry is the median over 315 held-out groups. A rollout cannot be split across GPUs, so one rollout bounds the simulation latency from below.}
  \label{tab:timing}
  \begin{tabular}{@{}lcc@{}}
    \toprule & Simulation & \Ltok{} \\
    \midrule
    Work per decision & 8 rollouts + scoring & 1 forward + filter \\
    Time, one worker & $\approx 891$\,s & 0.75\,ms \\
    Floor, unlimited GPUs & $\approx 111$\,s & 0.75\,ms \\
    \bottomrule
  \end{tabular}
\end{table}

\Cref{tab:timing} compares the measured cost of one \Ltok{} decision with the cost of evaluating the same eight candidates by closed-loop simulation.
The complete selection path, from trajectory normalization through the forward pass, the two-risk filter, and the argmax, takes 0.75\,ms per eight-candidate group on one GPU, with a tail latency of 0.77\,ms, and 4.53\,ms on one CPU thread, and it reproduces the frozen predictions and selections exactly.
Its scene input is one 4096-dimensional vector, compared with $3086\times4096$ values for \Lfull{} at the same parameter count.
Simulating the same eight candidates takes about 891\,s serially, and even with unlimited parallel hardware its latency stays above one rollout, about 111\,s.
This is a cost comparison, not an equivalence of quality: the simulator returns the executed outcome and the selector a learned prediction. It excludes candidate generation and the capture of $h$, and the simulation timings come from our offline acquisition pipeline rather than an optimized production simulator.

\section{Conclusion}
\label{sec:discussion}

We presented \headname{}, a risk-aware trajectory selector that reads the last hidden token of a frozen driving VLA's prompt prefill, filters candidates on predicted collision and offroad risk, and adds neither VLA training nor inference-time rollouts.
On a fixed pool of 525 scenes with eight candidates per group, it obtains lower selected regret, higher top-1 agreement, and a lower collision-at-fault or offroad rate than both Alpamayo's default first-sample rule and uniform random selection.
Matched ablations on the same pool show that a scene-blind input raises the event rate, whereas reading the full prefill sequence instead of its last token yields no supported gain in regret or top-1 agreement.

The selection rule was adopted on previously evaluated scenes, and the comparisons rely on pointwise intervals without multiple-comparison correction, so the results are exploratory rather than confirmatory. Risk filtering can sacrifice official-score performance, and non-dominated predictions do not guarantee safety. The results also rest on cached closed-loop outcomes of fixed candidates and do not cover repeated online replanning.
In future work, we will evaluate the selector under online replanning and explore whether the last-token representation, paired with a candidate trajectory, can seed a rollout in the style of a world model.

\bibliographystyle{IEEEtranN}
\bibliography{main}

\end{document}